\documentclass[letterpaper, 10 pt, conference]{ieeeconf}  

\IEEEoverridecommandlockouts                              

\usepackage{hyperref}
\usepackage{graphics} 
\usepackage{amsmath} 
\usepackage{amssymb}  
\usepackage{dsfont}
\usepackage{cleveref}
\usepackage{graphicx}
\usepackage{booktabs}
\usepackage{bbm}

\title{\LARGE \bf 
Learning from Pixels with Expert Observations
}

\author{Minh-Huy Hoang$^{1*}$, Long Dinh$^{2*}$, Hai Nguyen$^{3\dagger}$%
\thanks{$^{*}$Equal contribution,$^{1}$ University of Science, Ho Chi Minh City, Vietnam $^{2}$Hanoi University of Science \& Technology, Hanoi, Vietnam. $^{3}$Northeastern University, Boston, MA 02115, USA. $^\dagger$Corresponding author \texttt{nguyen.hai1@northeastern.edu}.}
}

\begin{document}

\maketitle
\thispagestyle{empty}
\pagestyle{empty}

\begin{abstract}
In reinforcement learning (RL), sparse rewards can present a significant challenge. Fortunately, expert actions can be utilized to overcome this issue. However, acquiring explicit expert actions can be costly, and expert observations are often more readily available. This paper presents a new approach that uses expert observations for learning in robot manipulation tasks with sparse rewards from pixel observations. Specifically, our technique involves using expert observations as intermediate visual goals for a goal-conditioned RL agent, enabling it to complete a task by successively reaching a series of goals. We demonstrate the efficacy of our method in five challenging block construction tasks in simulation and show that when combined with two state-of-the-art agents, our approach can significantly improve their performance while requiring 4-20 times fewer expert actions during training. Moreover, our method is also superior to a hierarchical baseline.
\end{abstract}

\section{INTRODUCTION}
\begin{figure}[htbp]
    \centering
    \includegraphics[width=0.85\linewidth]{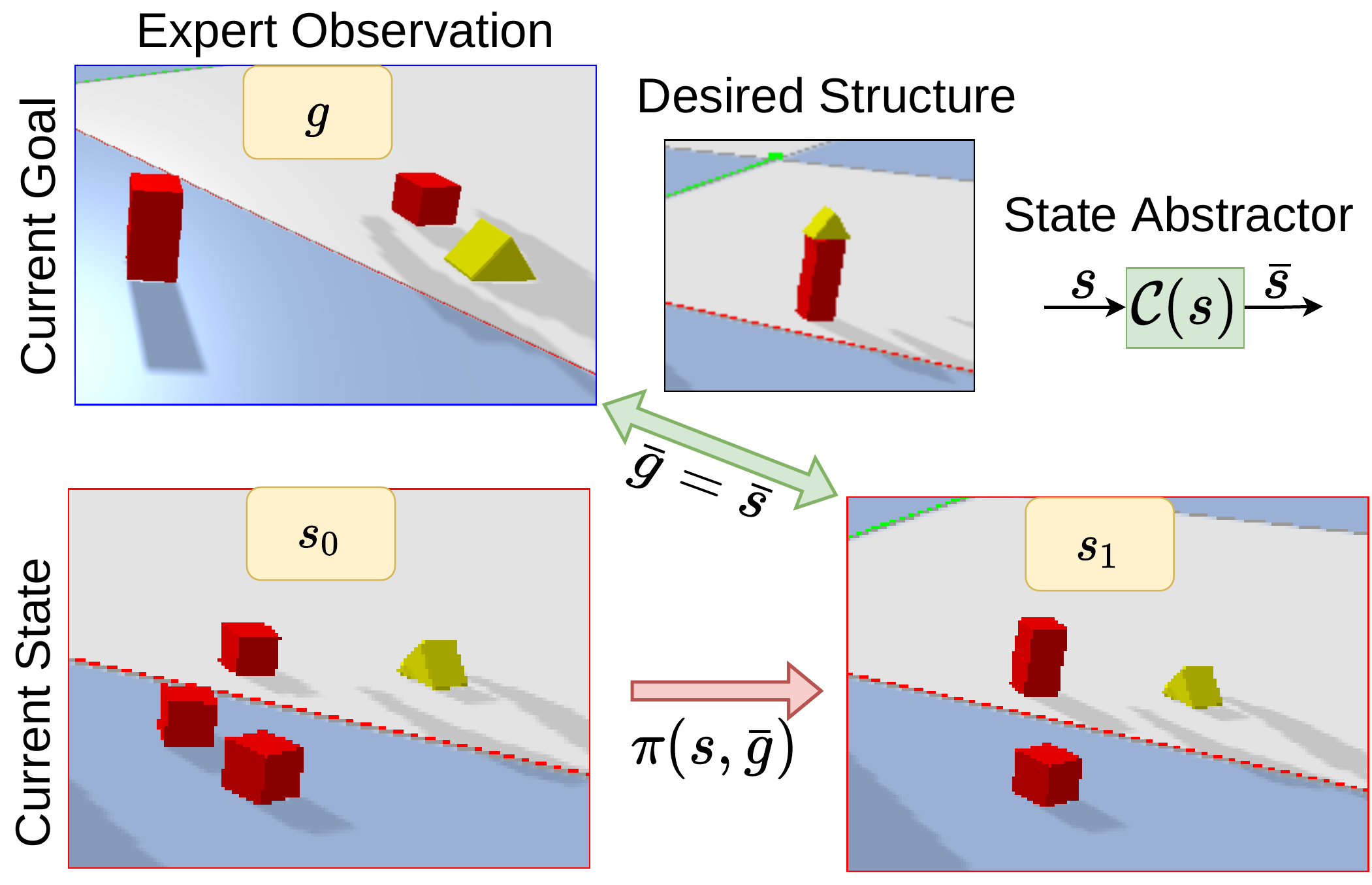}
    \caption{We utilize expert observations as intermediate visual goals to train a goal-conditioned policy. Using our approach, achieving a hard task (\emph{e.g.}, building a structure) equals achieving a series of easier goals sequentially.}
    \label{fig:motivation}
    \vspace{-10pt}
\end{figure}

Learning from observations without explicit guidance from an expert is an essential learning mechanism. Often, humans lack precise knowledge of the actions taken by a human expert, but they can deduce these actions by attempting to reproduce the expert's observations. For example, in \Cref{fig:motivation}, the task is to construct a desired structure consisting of a triangle on top of a stack of three blocks. At the current state $s_0$ when all blocks are on the ground, the optimal move is to form a stack of two blocks to reach the next state $s_1$, which closely resembles the scene an expert would observe at the subsequent step ($g$). Forming the stack should be much simpler than building the final structure, given only a pick and a place actions are needed. This example shows that an intermediate visual goal provided by expert observations can break down a complex task into more manageable subtasks and facilitate step-by-step task achievement.

This paper presents a hierarchical agent consisting of a \emph{fixed} top level and a \emph{learned} bottom level. At the top level, we use the indices of expert observations within an expert episode as the \emph{abstract goals} $\bar g$ (see~\Cref{fig:motivation}) for the bottom level to achieve. The bottom level implements a goal-conditioned policy $\pi (s, \bar g)$ that gradually realizes these abstract goals until the task is completed. The goal achievement is determined by comparing the \emph{abstract state} $\bar s$ produced by a \emph{state abstractor} $\mathcal{C}(s)$ with the corresponding abstract goal $\bar g$. The state abstractor is a multi-class classifier pre-trained using supervised learning with expert transitions.

This paper proposes a novel approach to improve the performance of current state-of-the-art methods in the setting of block construction tasks with sparse rewards and pixel-based observations. Specifically, we apply our approach to a version of Deep Q-Network (DQN)~\cite{mnih2015human} and Strict DQN from Demonstrations (SDQfD)~\cite{wang2020policy} and demonstrate substantial performance improvements in five challenging tasks. Our approach relies on fewer expert actions and utilizes expert observations as intermediate visual goals to create a simple yet effective learning strategy. Moreover, we also encode the domain symmetries into the state abstractor and the RL agent to increase the performance further. Our results indicate that our method outperforms the baselines, including a strengthened version of h-DQN~\cite{kulkarni2016hierarchical}, with 4-20 times fewer expert actions. Additional information including videos and code be found at \url{https://sites.google.com/view/leo-rb}.

\section{RELATED WORK}

\noindent \textbf{Goal-conditioned Reinforcement Learning} is a popular paradigm in which an RL agent seeks to achieve a goal, which may be a specific state (\emph{e.g.}, moving a block to a specific position) or an abstract concept (\emph{e.g.}, building a desired structure as in~\Cref{fig:motivation}). For more efficiency, goals are often diversified. Most notably, Hindsight Experience Replay~\cite{andrychowicz2017hindsight} (HER) \emph{relabels} previously visited states as goals. Instead of choosing goals randomly from past states like HER,~\cite{nguyen2019hindsight} selects goals from previously visited states based on how close they are to the original goal. However, in many tasks such as the one in~\Cref{fig:motivation}, useful goals may be difficult to visit by random actions, making relabelling unhelpful. In this work, we proposed to use expert observations as goals. Recent work in this area includes~\cite{shah2021ving}, which uses visual goals to navigate the open world,~\cite{davchev2021wish}, which chooses goals from successful demonstration trajectories, and~\cite{klee2022graph}, which employs a hand-coded abstraction function and goal graph for selecting goals to learn a multi-task policy in block construction domains. Our approach differs because we automatically propose abstract goals (using a learned state abstractor, not a hand-coded one like~\cite{klee2022graph}) instead of explicit ones selected from previous states like~\cite{shah2021ving, davchev2021wish} and operate in a single-task setting unlike~\cite{klee2022graph}. Moreover, we encode domain symmetries in our agent to enhance efficiency.

\noindent \textbf{Hierarchical Reinforcement Learning (HRL)} can break down a long-horizon RL task into smaller subtasks, which are more manageable. HRL has demonstrated superior performance over non-hierarchical approaches in various tasks~\cite{kulkarni2016hierarchical, levy2017learning, nguyen2022hierarchical}. Furthermore,~\cite{xie2020deep,gupta2019relay} utilize imitation learning to improve the hierarchical subgoal selection process to achieve better performance. Our approach is similar to h-DQN~\cite{kulkarni2016hierarchical}, which involves concurrently training two DQN~\cite{mnih2015human} agents. However, instead of learning two levels in parallel, which often causes instability, \emph{we only learn the bottom level}. At the top level, we use expert observations as goals for the bottom level to achieve, which we found to be more effective.

\noindent \textbf{Equivariant Neural Networks} can encode chosen symmetries within their structures, greatly enhancing generalization and sample efficiency. The idea was introduced by G-Convolution~\cite{GECN} and Steerable Convolutional Neural Networks (CNN)~\cite{SteerableCNNs} and has since been utilized in computer vision~\cite{equiCV1,equiCV2} and reinforcement learning~\cite{equiRL1}. In particular, earlier works~\cite{wang2020policy, wang2022equivariant, zhu2022sample, simeonov2022neural} have successfully applied equivariant networks to robot manipulation. We also found that using equivariant networks for our RL agent and state abstractor can increase task performance.

\noindent \textbf{Block Construction Tasks.} In the context of these challenging tasks, unlike other methods, such as object-factored models used in~\cite{li2020towards} and~\cite{funk2022learn2assemble}, our approach learns directly from image observations and sparse rewards. Additionally,~\cite{nair2018visual, nasiriany2019planning} perform model-based planning in a learned latent space to tackle block construction tasks from pixels. Recently,~\cite{lee2021beyond} proposed a new manipulation benchmark involving picking and placing objects of complex geometries, adopted by several~\cite{lee2022spend,zhou2021manipulator}. However, in this work, we consider block construction tasks of regular shapes, taken from the BulletArm benchmark~\cite{wang2022bulletarm}.

\section{BACKGROUND}
\subsection{Goal-conditioned DQN}
A goal-conditioned Markov decision process (MDP) is defined by the tuple $(\mathcal{S}, \mathcal{A}, \mathcal{G}, T, R)$, where $\mathcal{S}$ is the state space, $\mathcal{A}$ is the \emph{discrete} action space, $\mathcal{G}$ is the goal space, $T(s,a,s') = p(s' \mid s,a)$ is the transition function that gives the probability of reaching state $s'$ given an action $a \in \mathcal{A}$ is taken in state $s$, $R (s, a, s', g)$ is the reward function. The objective is to find a goal-conditioned policy $\pi(s,g)$ that maximizes the expected discounted return $\mathbb{E}_{\pi}[\sum_{t=0}^{\infty} \gamma^t r_t \mid s_0, g]$ with a discount factor $\gamma \in [0, 1)$, starting with initial state $s_0$ and goal $g$. In contrast to DQN~\cite{mnih2015human}, which learns the optimal Q-function using transitions of the form $(s, a, s', r)$, the optimal goal-conditioned Q-function $Q^*(s, a, g)$ is learned using transitions of the form $(s, a, s', r, g)$~\cite{schaul2015universal}. From $Q^*(s, a, g)$, we can obtain an optimal goal-conditioned policy $\pi^* (s, g)$:
\begin{equation}
    \pi^* (s, g) = \operatorname{argmax}_a Q^*(s, a, g).
\end{equation}

\subsection{Group Equivariance and Invariance}
Given a symmetry group $G$ and an element $ \alpha \in G$, a function $f: X \rightarrow Y$ is \emph{equivariant} if $f(\alpha x) = \alpha f(x)$, and \emph{invariant} if $f(\alpha x)=f(x)$. More precisely, how $\alpha$ acts on $X$ or $Y$ depends on the \emph{representation} $\rho$ of the symmetry group $G$. For instance, if $G$ is a group of planar rotations and $x$ is a single-channel image, then $\alpha x = \rho(\alpha)x$ will be a multiplication between a rotation matrix $\rho(\alpha)$ and an image $x$, resulting in a rotated image. However, for clarity, in the remainder of the paper, we use $\alpha x = \rho(\alpha)x $ to denote the action of $\alpha$ on $x$ without referring to any specific representation $\rho$. For more details about group representations, please refer to~\cite{weiler2019general}.

\subsection{Group-invariant MDP in SE(2)}
\label{sec:group_invariant}
Many manipulation tasks of rigid blocks from top-down images~\cite{wang2022bulletarm} have the symmetry about the rotation and translation in plane $\mathbb{R}^{2}$ (\emph{i.e.}, spanning the special Euclidean group SE(2)). Such tasks can be defined as a group-invariant MDP~\cite{wang2022so}, which is invariant in SE(2) by satisfying the following conditions for all group element $\alpha \in \text{SE(2)}$:
\begin{itemize}
    \item Reward Invariance:
\begin{align}\label{eq:reward_invariance}
    R(s, a, s') = R(\alpha s, \alpha a, \alpha s').
\end{align}
    \item Transition Invariance:
\begin{align}\label{eq:transition_invariance}
    T(s, a, s') = T(\alpha s, \alpha a, \alpha s').
\end{align}
\end{itemize}
The key feature of a group-invariant MDP is that its optimal Q-function is invariant, \emph{i.e.}, $Q^*(s, a) = Q^*(\alpha s, \alpha a)$ for all $\alpha \in \text{SE(2)}$. This invariance property is the foundation to build very sample-efficient RL agents~\cite{wang2022equivariant, wang2020policy, wang2022so} by directly encoding the property into the Q-network structure.

\subsection{Strict DQfD (SDQfD)}
\label{sect:sdqfd}
Besides fitting the Q-function $Q(s, a)$ similar to DQN~\cite{mnih2015human}, DQfD~\cite{hester2018deep} leverages expert demonstrations by \emph{permanently} storing the expert transitions inside the replay buffer. Specifically, DQfD additionally minimizes a margin loss to force the value of non-expert actions to be below the value of expert actions given the same state:
\begin{equation}
    \mathcal{L}_{M} = \mathbb{E}_{s,a^{e}}[\max_{a}(Q(s,a) + l(a,a^{e})) - Q(s,a^{e})]\,,
\end{equation}
where $a^{e}$ is an expert action in state $s$ and $l(a,a^{e})$ is penalty that is positive when $a$ is non-expert and is 0 otherwise.

In \emph{large} action spaces, SDQfD~\cite{wang2020policy} utilizes a similar penalty but applies it to \emph{all} non-expert actions that have Q-values higher than that of an expert action (given the same state) minus the penalty. The replacement of $\mathcal{L}_{M}$ is a new \emph{strict} margin loss defined as: 
\begin{equation}\label{eq:sdqfd}
    \mathcal{L}_{SM} = \frac{1}{|A^{s,a^{e}}|}\sum_{a \in A^{s,a^{e}}}[Q(s,a) + l(a,a^{e}) - Q(s,a^{e})]\,,
\end{equation}
where $A^{s,a^{e}}$ is set of actions satisfied:
\begin{equation}
    A^{s,a^{e}}=\{a \in \mathcal{A} \mid  Q(s,a) + l(a,a^{e}) > Q(s,a^{e})\}.
\end{equation}
SDQfD considerably outperformed both DQfD and DQN with demonstrations in many block construction tasks with sparse rewards~\cite{wang2020policy}.

\section{STRUCTURE CONSTRUCTION FROM PIXELS}
\label{sect:task_intro}

\begin{figure}[htbp]
    \centering
    \includegraphics[width=0.95\linewidth]{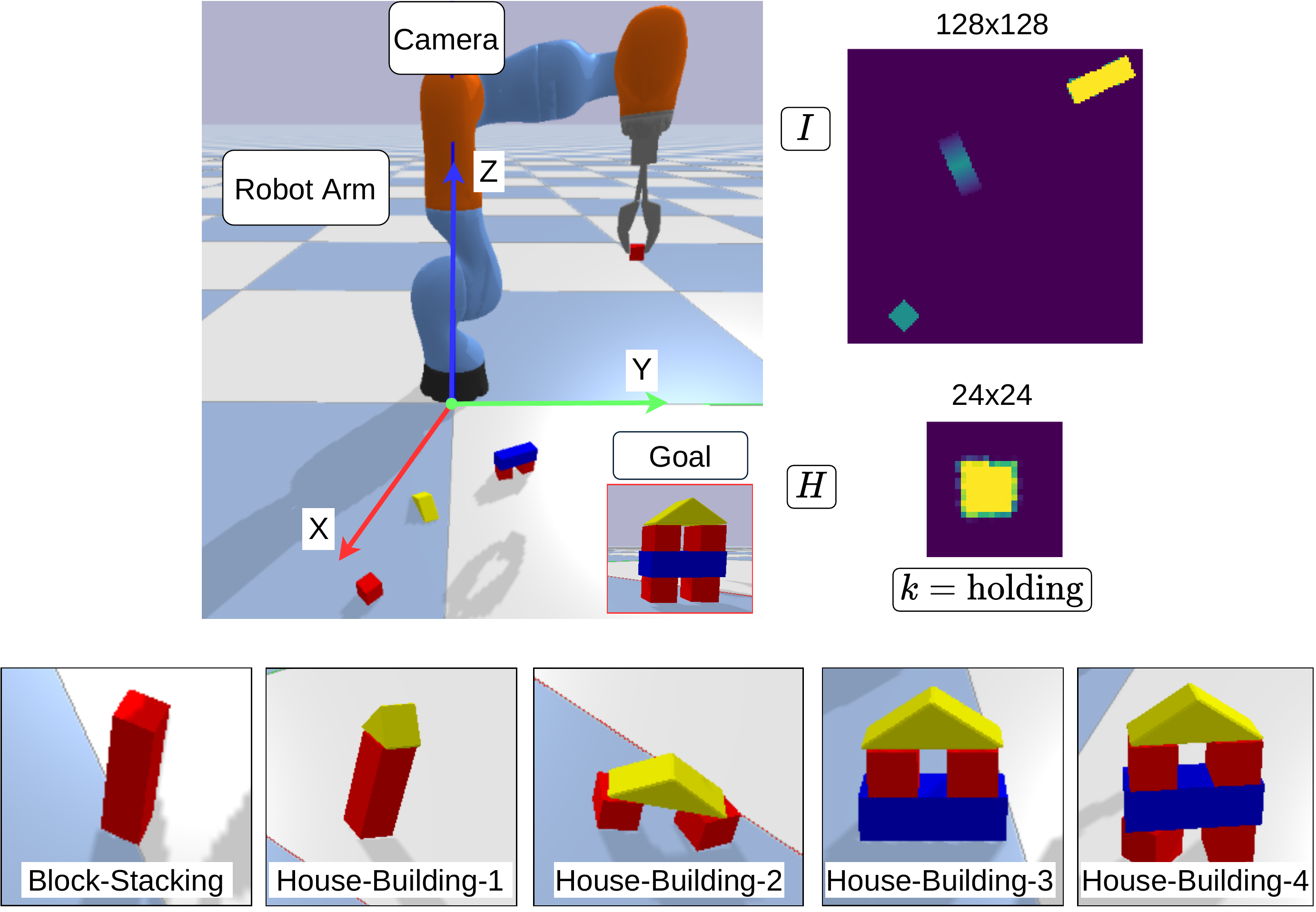}
    \caption{A visual description of our block construction tasks and a list of five tasks considered for experiments.}
    \label{fig:intro}
    \vspace{-10pt}
\end{figure}

\begin{table}[htbp]
    \caption{More Details on Tasks.}
    \centering
    \begin{tabular}{l c c c c c}
         \textbf{Task} & \texttt{BS} & \texttt{HB1} & \texttt{HB2} & \texttt{HB3} & \texttt{HB4} \\ \toprule
         Number of  Blocks & 4  & 4  & 3  & 4  & 6 \\ 
         Number of Optimal Steps & 6  & 6  & 4  & 6  & 10 \\ 
         Max Number of Steps   & 10 & 10 & 10 & 10 & 20 \\  
         \bottomrule
    \end{tabular}
    \label{tab:task_details}
\end{table}

\Cref{fig:intro} shows five structure construction tasks in the BulletArm benchmark~\cite{wang2022bulletarm}, namely Block-Stacking (\texttt{BS}) and House-Building-$i$ (\texttt{HBi}) with $i=\{1, 2, 3, 4\}$. In these tasks, a robot arm must construct a desired structure from unstructured blocks using top-down depth images of the scene taken by a camera on top (see an example task in~\Cref{fig:intro}). At the beginning of each episode, the agent is presented with necessary blocks with random poses to build the desired structure (see~\Cref{tab:task_details} for more task details).

\noindent \textbf{State Space.} We consider a state $s = (I, H, k)$, where $I \in \mathbb{R}^{128 \times 128}$ is a top-down depth image of the scene, $H \in \mathbb{R}^{24\times24}$ is an in-hand image denoting the current object in the gripper, and $k \in \{ \texttt{holding}, \texttt{empty} \}$ refers to the current grasping status of the gripper. The in-hand image $H$ is an orthographic projection of the partial point cloud where the last pick occurred (see~\Cref{fig:intro}).

\noindent \textbf{Action Space.} An action $a = (p, x, y, \theta)$ where $p \in \{\texttt{pick}, \texttt{place}\}$, $(x, y)$ are the pixel coordinate in $I$ that the agent wants to perform a pick or a place, and $\theta \in \{ 0, \frac{\pi}{16}, \frac{2\pi}{16}, \dots, \frac{31\pi}{16}\}$ is the discretized rotation of the gripper around the $z$ axis. The value of $p$ decides whether to close the fingers (pick) or open the fingers (place). In this setting, the action space is \emph{large}, \emph{e.g.}, there are $2 \times 128\times128\times32 = 1,048,576$ possible discrete actions.

\noindent \textbf{Reward.} The agent is given a sparse reward $r = 1.0$ if accomplishing the desired structure, else $r = 0.0$.

\noindent \textbf{Translation and Rotation in SE(2). } Given an arbitrary rotation and translation in the plane $\alpha \in \text{SE(2)}$, here we define how it would operate on a state $s$ and an action $a$. First, $\alpha$ operates on $s = (I, H, k)$ by translating and rotating $I$ but leaving $H$ and $k$ unchanged:
\begin{align}
    \alpha  s = (\alpha  I, H, k).
\end{align}
For an action $a = (p, x, y, \theta)$, it rotates and translates the spatial components $(x, y, \theta) \in \text{SE(2)}$ but leaving $p$ unchanged:
\begin{align}
    \alpha  a = (p, \alpha  x, \alpha y, \alpha \theta).
\end{align}

\begin{figure}[htbp]
    \centering
    \includegraphics[width=0.95\linewidth]{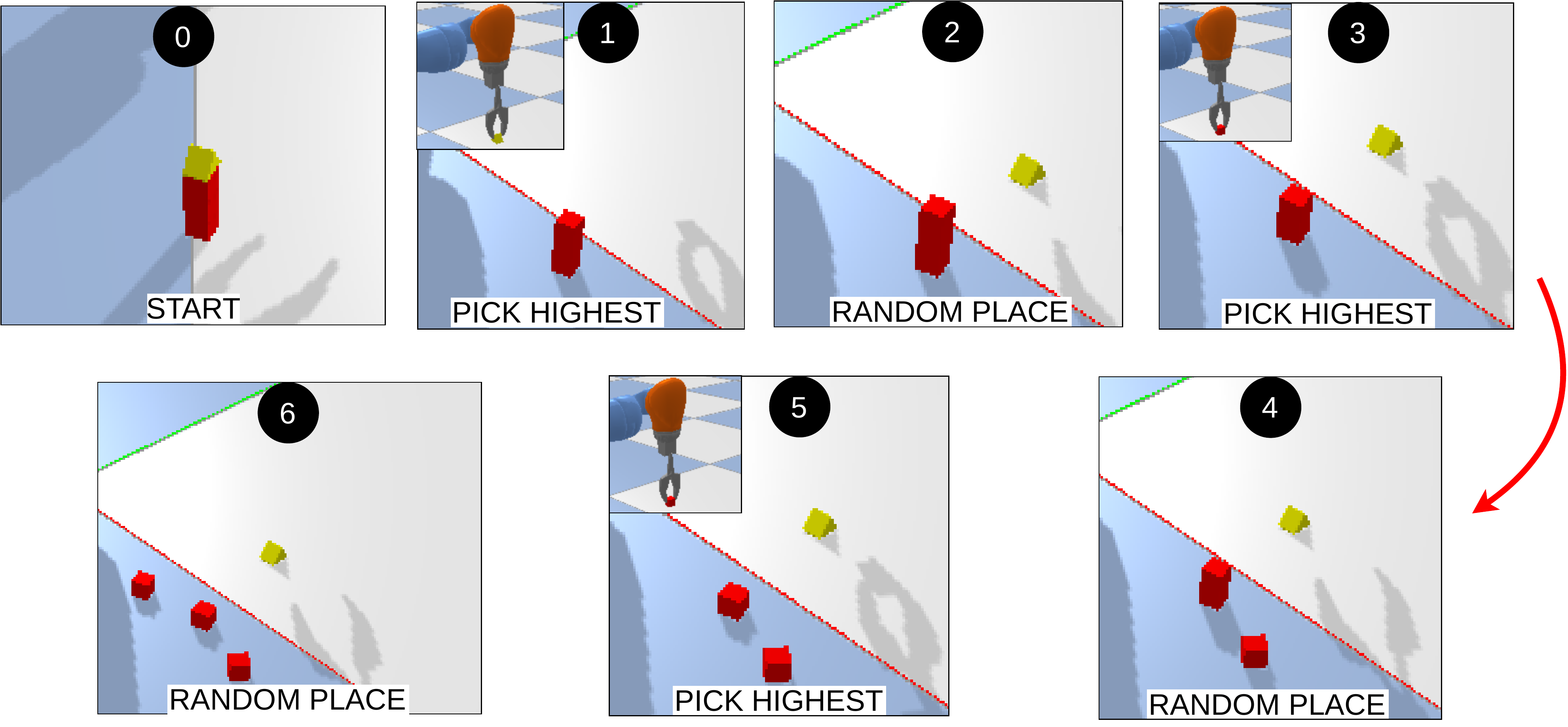}
    \caption{$0 \rightarrow 6$: A deconstruction planner decomposes a fully built structure by randomly picking the highest block and placing it on the ground until the structure is fully decomposed. Reverting $6 \rightarrow 0$ results in a demonstration episode. Besides indexing, numbers are also used for abstract states.}
    \label{fig:deconstruct}
\end{figure}

\noindent \textbf{Generating Expert Demonstrations.} To collect expert demonstrations, we utilize a simple deconstruction planner similar to~\cite{wang2022bulletarm, wang2020policy}, illustrated in~\Cref{fig:deconstruct}. The process starts with the desired structure fully built; then, the highest object is picked and randomly placed on the ground until the structure is fully decomposed. We then reverse the trajectory to obtain an expert episode. This process allows us to efficiently generate expert episodes without requiring task-specific expert policies, which would be difficult to obtain.

\section{LEARNING WITH EXPERT OBSERVATIONS}
\subsection{Formulating as a Goal-Conditioned MDP}
We formulate the problem of building a structure with $N$ blocks as a goal-conditioned MDP $M_G = (\mathcal{S}, \mathcal{A}, \bar{\mathcal{G}}, T, R_g)$, where $\mathcal{S}$, $\mathcal{A}$, and $T$ are the same as in the original problem. The differences are the goal space and the reward function:
\begin{align}
    \bar {\mathcal{G}} &= \{0, 1,  \dots,  2N-2 \} \\
    R_g(s, a, s', \bar g) &=  \begin{cases}
          r,  & \text{if } \mathds{1}(\bar {s'}, \bar g) = 1 \\
          r - 1,  & \text{otherwise} \,,
          \end{cases}
\end{align}
where $\bar {\mathcal{G}}$ is an abstract goal space, $\bar g \in \bar {\mathcal{G}}$ is an abstract goal, $\mathds{1}(.)$ is the indicator function, $r$ is is the environment reward (see~\Cref{sect:task_intro}), and $\bar{s'}$ is the abstract state of $s'$. Notice that besides rewarding $0$ when the abstract goal is achieved and $-1$ otherwise, our reward function gives the agent more reward when the final structure is achieved successfully, \emph{i.e.}, when $r = 1.0$. This additional reward is to emphasize the importance of achieving the desired structure. Moreover, the indices in $\bar {\mathcal{G}}$ are exactly the indices of expert observations for the task (see~\Cref{fig:deconstruct} for the task \texttt{HB1}).

\begin{figure}[htbp]
    \centering
    \includegraphics[width=0.95\linewidth]{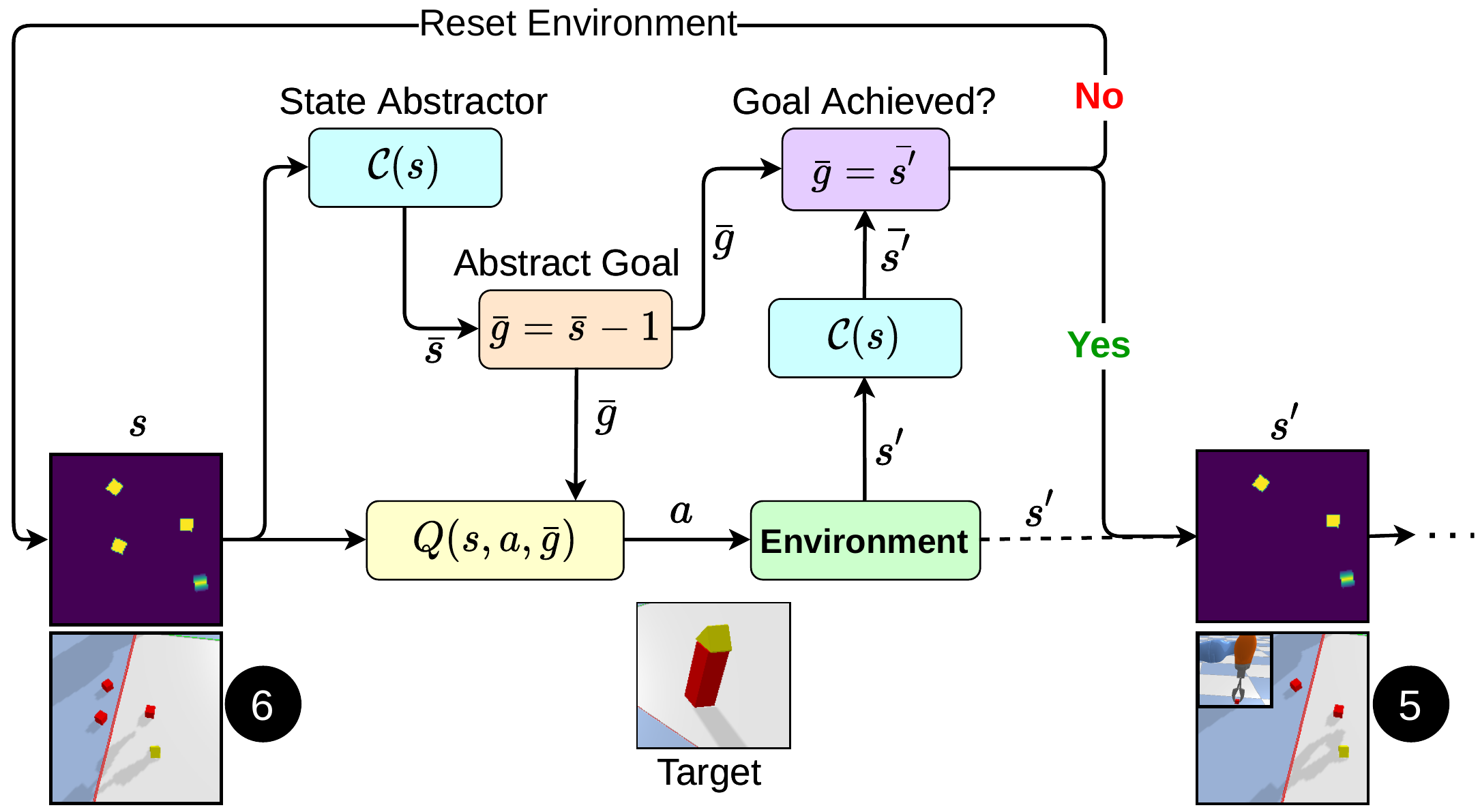}
    \caption{The agent starts at a state $s$ where all the blocks are on the ground. In this state, the abstract state is $\bar s = 6$, as illustrated in Figure~\ref{fig:deconstruct}. To construct the desired structure in the middle, the agent must transition to the next state, $s'$, which should resemble the scene numbered 5 in Figure~\ref{fig:deconstruct}, where the agent picks up a red block. Therefore, the abstract goal is $\bar g = \bar {s'} = \bar s - 1$. If the agent succeeds in achieving this goal in the following step, it will be supplied with another goal. Otherwise, the environment will be reset.}
    \label{fig:overview}
\end{figure}

To make the idea more concrete, \Cref{fig:overview} shows how our agent tackles the \texttt{HB1} task. Initially, the agent is in state $s$, where all the blocks are on the ground. This state corresponds to an abstract state $\bar s = 6$, as depicted in~\Cref{fig:deconstruct}. To build the desired structure, the agent must transition to the next state $s'$ where it picks up a red block. This next state should resemble the scene labeled as number 5 in~\Cref{fig:deconstruct}, making the abstract goal $\bar g = \bar {s'} = \bar s - 1 = 5$. Since $s$ and the desired $s'$ are only one pick action away in~\Cref{fig:deconstruct}, the agent should only need a single timestep to reach $s'$ from $s$. Therefore, if the agent achieves the abstract goal in the next timestep, it is presented with another goal; otherwise, we reset the episode. In the ideal case, the process continues until the desired structure is fully built, \emph{i.e.}, when the abstract goal $\bar{g} = 0$ is achieved.

To facilitate such a learning process, our agent comprises the following components:

\begin{itemize}
    \item A state abstractor $\mathcal{C}(s)$:
    \begin{equation}
     \mathcal{C}(s): \mathcal{S} \rightarrow \bar{\mathcal{S}}\,,
    \end{equation}
    where $\bar{\mathcal{S}} \equiv \bar{\mathcal{G}}$.
    \item An \emph{implicit} goal-conditioned policy extracted from a Q-function:
    \begin{equation}
    \pi (s, \bar g) = \operatorname{argmax}_a Q(s, a, \bar g). 
    \end{equation}
\end{itemize}
We also note that our formulation can be easily extended to accommodate a more general goal space $\mathcal{G}$. In such cases, the goals can be actual expert observations, such as images, rather than abstract states representing those observations.

\begin{figure*}[htbp]
    \centering
    \includegraphics[width=0.8\linewidth]{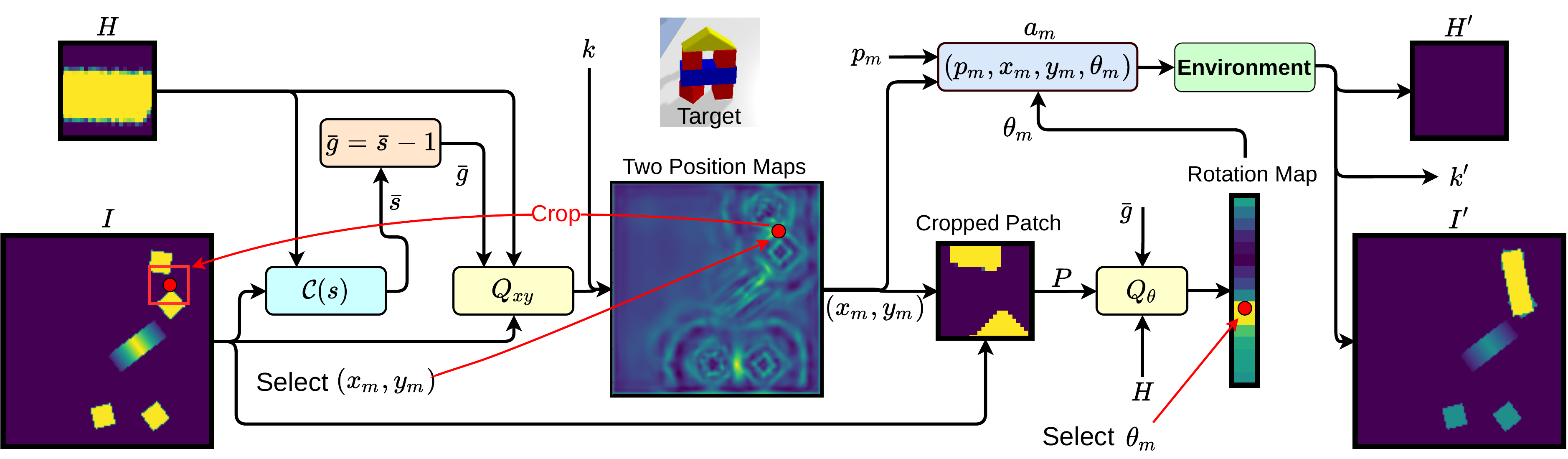}
    \caption{Learning $Q(s, a, \bar g)$ in \texttt{HB4} using goal-conditioned ASR~\cite{wang2020policy}. Learning $Q(s, a, \bar g)$ is divided into learning a position Q-function $Q_{xy}$ and a rotation Q-function $Q_\theta$. Picking or placing ($p_m$) is chosen based on the grasping status $k$. The red dots denote the selected pixel location to pick/place $(x_m, y_m)$ using $Q_{xy}$ and the subsequently selected rotation $\theta_m$ using $Q_\theta$. $Q_\theta$ decides $\theta_m$ using a $H$ and a small patch $P$ around $(x_m, y_m)$ of $I$.}
    \label{fig:ASR}
    \vspace{-10pt}
\end{figure*}

\subsection{State Abstractor}
\label{sec:state_abstractor}

$\mathcal{C}(s)$ can be considered as a multi-class classifier that maps each state $s$ to its corresponding abstract state $\bar{s}$. It is responsible for determining goal achievement for our agent. To train it, we use a dataset $\mathcal{D}_{train}$ of pairs $(s, \bar s)$ collected from expert demonstrations. As shown in~\Cref{fig:deconstruct}, for each state $s$ in the expert episode, we use the index of $s$ inside the episode to be $\bar s$. Given $\mathcal{D}_{train}$, $\mathcal{C}(s)$ is trained by minimizing the following cross-entropy loss:
\begin{equation}
    \mathcal{L}_{CE} = -\frac{1}{|\mathcal{D}_{train}|}\sum_{i=1}^{|\mathcal{D}_{train}|}\sum_{j=0}^{2N-2} \mathds{1}[\bar{s}_i=j]\log \mathcal{C}(s_i)_j \,,
\end{equation}
where $|\mathcal{D}_{train}|$ denotes the size of $\mathcal{D}_{train}$. For each given task, we pre-train $\mathcal{C}(s)$ and \emph{freeze} it to use for checking goal achievement. In addition, given a group element $\alpha \in \text{SE(2)}$, then $\mathcal{C}(s) = \mathcal{C} (\alpha s)$, \emph{i.e.}, the abstract state should be the same if we rotate and translate the state (\emph{e.g.}, top-down image). Therefore, we construct it as an invariant function.

\subsection{Learning $Q(s, a, \bar g)$}
Augmented state representation (ASR)~\cite{wang2020policy} is a powerful method for learning the Q-function in domains with large action spaces, as in our tasks. The idea is to transform an MDP with a large action space into a new one with more states but fewer actions. We also adopt ASR to learn $Q(s, a, \bar g)$ but modify it for our goal-conditioned MDP.

\subsubsection{Transition Division}
We first divide the action into three components: the \emph{pick/place} component: $p$, the \emph{position} component: $a_{xy} = (x, y)$, and the \emph{rotation} component: $\theta$. Now considering that the agent wants to achieve an abstract goal $\bar g$ at state $s$ by deciding the next action $a_m = (p_m, x_m, y_m, \theta_m)$. Depending on the grasping status $k$, whether to pick or place $p_m$ can be chosen using simple logic: we pick when the gripper is currently empty and place when it is holding something, \emph{i.e.},
\begin{equation}
    p_m = \begin{cases}
          \texttt{pick},  & \text{if } k = \texttt{empty} \\
          \texttt{place},  & \text{if } k = \texttt{holding}
          \end{cases}
\end{equation}
For selecting $(x_m, y_m, \theta_m)$, we need to learn $Q(s, a, \bar g)$. ASR does that by dividing the \emph{original} transition, which is
\begin{equation}
    s, \bar g \xrightarrow {a} s' \,,
\end{equation}
into two transitions. The position transition only involves $a_{xy}$:
\begin{equation}\label{eq:xy}
    s, \bar g \xrightarrow{a_{xy}} \tilde{s} \,,
\end{equation}
where $\tilde{s} \in \mathcal{S}$ can be considered as the resulting next state of the original transition after taking action $a = (p, a_{xy}, 0)$. The rotation transition only involves $\theta$:
\begin{equation}\label{eq:theta}
    \tilde{s}, \bar g \xrightarrow{\theta} s'.
\end{equation}

\subsubsection{Learning Factorial Q-functions}
 By the transition division, learning $Q(s, a, \bar g)$ can be effectively performed by learning a position Q-function $Q_{xy}(s, a_{xy}, \bar g)$ and a rotation Q-function $Q_{\theta}(\tilde s,  \theta, \bar g)$.
 \begin{itemize}
     \item $Q_{xy}$ decides where we should perform pick/place in the image. As shown in~\Cref{fig:ASR}, it outputs \emph{two} position maps with the same size as $I$'s, corresponding for $p_m = \texttt{pick}$ and $p_m = \texttt{place}$. These two maps represent the Q-values of picking or placing at a certain pixel position concerning the current abstract goal $\bar g$. 
     Now, during exploitation, the position to perform a pick/place can be chosen greedily:
     \begin{equation}
     (x_m, y_m) = \operatorname{argmax}_{a_{xy}}Q_{xy}(s, a_{xy}, \bar g).
     \end{equation}
     \item  $Q_\theta$ decides the rotation $\theta_m$ of the gripper when we have decided where to pick/place. To do that, ASR only considers a small patch $P$ of $I$, centered around $(x_m, y_m)$. Instead of learning the original $Q_\theta$, we learn $Q_\theta(H, P, \theta, \bar g)$, which should be sufficient to decide the best $\theta_m$ given the information contained in $H$ and $P$. During exploitation, we simply greedily select the orientation using a rotation map outputted by $Q_\theta$:
     \begin{equation}
         \theta_m = \operatorname{argmax}_{\theta}Q_\theta(H,P, \theta, \bar g).
     \end{equation}
 \end{itemize}

\subsubsection{Invariant Factorial Q-functions}
We show that $Q_{xy}$ and $Q_\theta$ are invariant so that they can be constructed as invariant models. Following~\Cref{sec:group_invariant}, we will show that the two Q-functions correspond to group-invariant MDPs with invariant reward and transition functions.

From~\Cref{eq:xy}, we can observe that the Q-function $Q_{xy}(s, a_{xy}, \bar g)$ corresponds to the original goal-conditioned MDP $M_G$ except that the action space is constrained by only allowing $\theta = 0$. We convert $M_G$ to be a traditional MDP by using an augmented state $(s, \bar g)$, which changes the reward function of $M_G$ into $R'_g((s, \bar g), a_{xy}, (s', \bar g))$. The new reward function is invariant because achieving $\bar g$ should be independent of the position and orientation of the scene. Moreover, the transition is invariant because, in the task setting, the outcome of an action should be invariant when the scene and the action rotate and translate with the same amount. Therefore, $Q_{xy}$ is invariant.

From~\Cref{eq:theta}, because $\tilde{s} \in \mathcal{S}$, the orientation transition $\tilde{s}, \bar g\xrightarrow{\theta} s'$ correspond to a transition in $M_G$ except that now with actions being the rotations $\theta$ only. Therefore, the same above logic can apply here with an augmented state $(\tilde{s}, \bar g)$ to show that $Q_\theta$ is also invariant.

\begin{figure*}[htbp]
    \centering
    \includegraphics[width=1.0\linewidth]{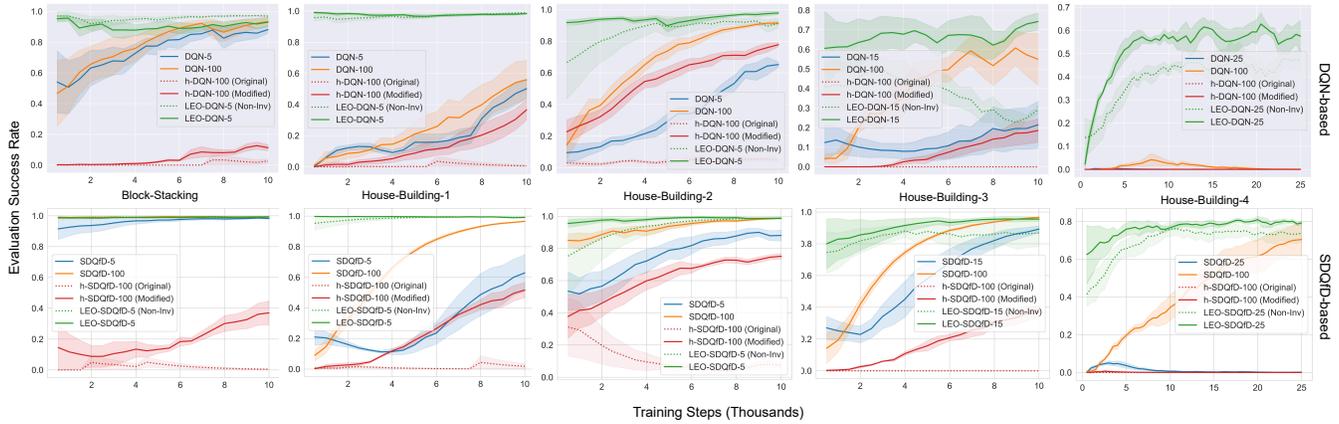}
    \caption{Evaluation success rate averaged over four seeds with shaded one standard error. The top row (grey background) contains DQN-based agents, and the bottom row (white background) contains SDQfD-based agents. Note that all agents utilize equivariant ASR~\cite{wang2022equivariant}.}
    \label{fig:exp_result}
    \vspace{-10pt}
\end{figure*}

\section{EXPERIMENTS AND RESULTS}
In the five structures considered, we want to investigate whether our approach of learning from expert observations (LEO) can help improve the performance of existing best RL agents~\cite{wang2022bulletarm} for our tasks, namely, DQN and SDQfD.
\subsection{Agents}
To investigate the possible benefits of our approach on DQN and SDQfD, we consider the following agents:
\begin{itemize}
    \item \underline{DQN/SDQfD-$x$}~\cite{wang2022bulletarm}: These are baselines that use $x$ demonstration episodes (\emph{with expert actions}) by permanently storing them in the replay buffer. Note that SDQfD better uses the demonstration using the margin loss (see~\Cref{eq:sdqfd}). Both employ equivariant ASR~\cite{wang2022equivariant} and are currently state-of-the-art agents for our tasks (see~\cite{wang2022bulletarm}).
    \item \underline{LEO-DQN/SDQfD-$x$} (Ours): The versions of DQN/SDQfD-$x$ that use our approach (LEO-). We also benchmark LEO-based agents' performance using non-invariant state abstractors (\underline{Non-Inv}).
    \item \underline{h-DQN-100}, \underline{h-SDQfD-100} are two-level hierarchical agents based on h-DQN~\cite{kulkarni2016hierarchical} agent that learn both levels concurrently. The top learns to produce abstract goals, and the bottom attempts to achieve them. Like our agent, the bottom uses the same pre-trained $\mathcal{C}(s)$ to determine goal achievement. We consider two versions of these agents. The original version (\underline{Original}) follows the original reward scheme by only using the environment rewards for the top level. We improve the original agents (\underline{Modified}) by additionally rewarding the top with $0$ and $-1$ if the bottom can or cannot achieve its abstract goals. Furthermore, to stabilize training, we follow the practice in~\cite{kulkarni2016hierarchical} to pre-train the bottom level with random abstract goals before training the two levels in parallel. It is also worth noting that these agents also use equivariant ASR and $100$ demonstration episodes.
\end{itemize}

\subsection{Evaluation Metrics}
We compare methods using the \emph{success rates of an evaluation agent}, which is evaluated for every 500 training (environment) steps. We train all agents in 25,000 environment steps in \texttt{HB4} (the hardest task) and in 10,000 environment steps in the four remaining structures. The reported results are averages of 4 seeds with shaded one standard error.

\subsection{Learning Curves}
We visualize all learning curves in~\Cref{fig:exp_result}, where the top row shows DQN-based agents and the bottom shows SDQfD-based agents. Given the same demonstration episodes, LEO-based agents outperform the original agents regarding final performance and learning speed. The superiority is consistent over all tasks and all algorithms (DQN/SDQfD) tested. Moreover, given more ($100$) expert episodes, the original and hierarchical agents are still outperformed by our agents with 4-20 times fewer expert actions used, \emph{e.g.}, our agents only use 5, 15, or 25 expert episodes. The figure also indicates that using an invariant state abstractor (solid green lines) benefits our agents more than not (dotted green lines) in all cases (except for DQN-based agents in \texttt{BS}), regardless of the algorithms used. Finally, the original versions of h-DQN and h-SDQfD perform much worse than our modified version. They fail to learn in all tasks, probably due to a lack of positive environment rewards for the top level to learn. In contrast, with the modified reward function by rewarding/penalizing the top when it produces achieved/unachieved abstract goals, the modified version performs much better in all tasks.

\subsection{Learning with No Expert Actions}
\begin{figure}[htbp]
    \centering
    \includegraphics[width=0.95\linewidth]{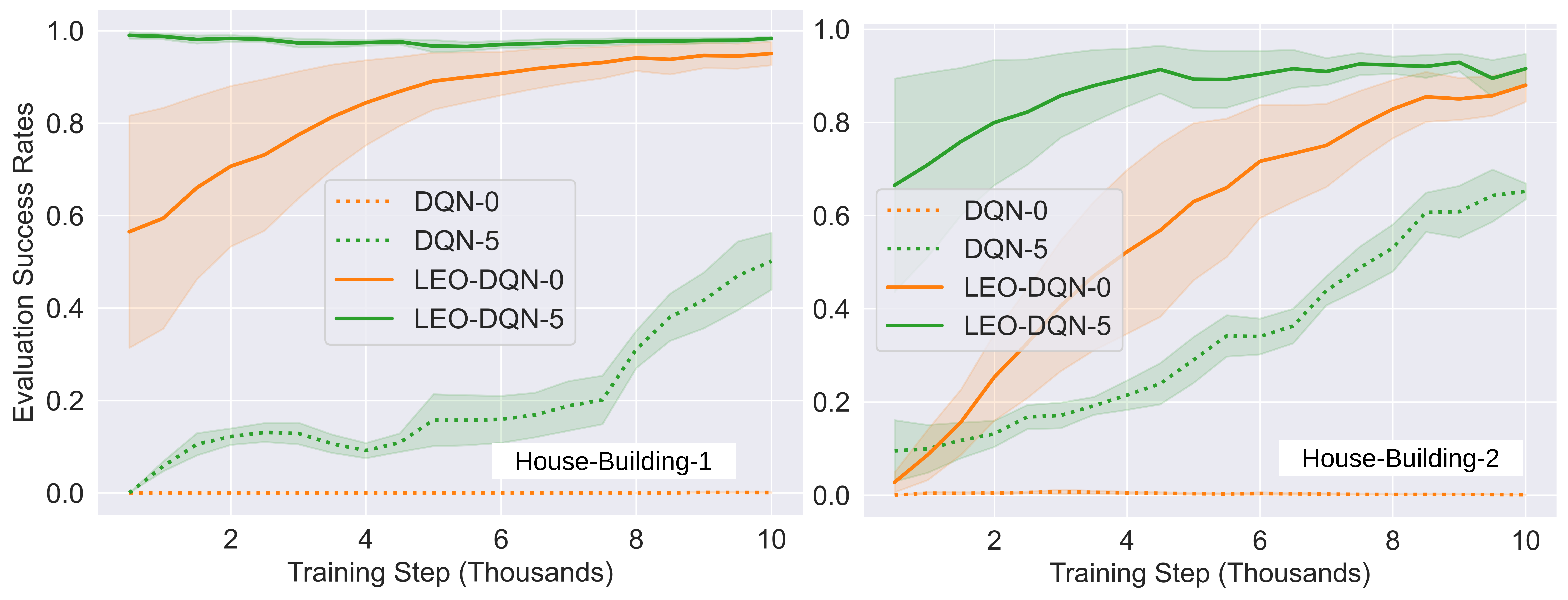}
    \caption{Performance in \texttt{HB1} and \texttt{HB2} when no expert actions are used.}
    \label{fig:no_expert}
    \vspace{-15pt}
\end{figure}

We hypothesize that our approach can benefit learning in specific tasks even when no expert actions are given. To verify this hypothesis, we compare DQN and LEO-DQN in \texttt{HB1} and \texttt{HB2} with \emph{no} expert actions (we exclude SDQfD because expert actions are required to implement it, see~\Cref{eq:sdqfd}).~\Cref{fig:no_expert} shows that LEO-DQN-0 agents (solid orange), even though they perform worse than LEO-DQN-5 (as expected with fewer expert episodes), can still greatly outperform DQN-0 agents (dotted orange lines). Specifically, with no expert actions provided, DQN-0 agents fail to learn both tasks, while LEO-DQN-5 can nearly master the two tasks at the end of training.

\subsection{Implementation Details}
Our code is implemented in PyTorch~\cite{paszke2019pytorch}, and all optimization uses the Adam~\cite{kingma2014adam} optimizer. Below we only highlight key implementation for Q-functions and the state abstractor (see our website for their detailed diagrams).

\noindent \textbf{Discrete Approximation of SE(2).} In practice, it is common to approximate the infinite group SE(2) with finite groups~\cite{weiler2019general, cohen2016steerable}. Follow the previous work~\cite{wang2020policy}, we approximate SE(2) with $\hat{\text{SE}}(\text{2})$, which is the cross product between $\{1, \dots, 128\} \times \{1, \dots, 128\} \subset \mathbb{Z}^2$ and the cyclic group $C_{32} = \{ k\pi/16 : 0 \leq k < 32, k \in \mathbb{Z}\}$.

\noindent \textbf{Q-functions} of all agents are constructed by equivariant ASR networks~\cite{wang2020policy}, implemented using the \texttt{e2cnn}~\cite{weiler2019general} library. Non-equivariant inputs, such as in-hand images and abstract goals, are integrated into our agents using Dynamic Filter~\cite{wang2022equivariant}. We use the dihedral group $D_{4}$ for $Q_{xy}$ and $C_{32}$ group for $Q_\theta$, where $D_{4}$ is $C_{4}  = \{ k\pi/2 : 0 \leq k < 4, k \in \mathbb{Z}\}$ combined with the reflection symmetry. All agents are trained with a learning rate of 1e-4. The batch size is set to 32, and the discount is 0.95. We run five simulated environments in parallel to collect transitions for training. The size of the cropped patch $P$ is $24 \times 24$ pixels.

\noindent \textbf{State Abstractor} uses two branches to extract features from the top-down and the in-hand images. The branch for top-down images is a series of five [Conv+Max-Pool] modules. The branch for in-hand images only consists of three such modules. The extracted features are then concatenated before going through three fully-connected layers and a softmax layer at the end. The invariant state abstractor replaces the traditional CNNs with equivariant CNNs using the group $D_{4}$. For each task, we train our state abstractors in 12,000 training steps with a batch size of 32 using a learning rate of 1e-3. The number of samples per class for training $\mathcal{C}(s)$ in \texttt{BS} and \texttt{HB1} is 250, in \texttt{HB2} and \texttt{HB3} is 500, and in \texttt{HB4} is 1000.

\section{CONCLUSIONS}

This paper showed that expert observations could be beneficial learning signals when expert actions are only sparingly provided. Even though we only demonstrate the improved performance on DQN and SDQfD, we project that our approach should benefit other off-policy learning algorithms. The limitation of our approach is that the state abstractor, which acts similarly to human eyes, must be sufficiently good, ideally trained from \emph{optimal} experts. Moreover, while we have used equivariant neural networks to improve the efficiency and generalization of our state abstractor, it cannot currently handle outliers~\cite{yang2021generalized}, which can be addressed using approaches~\cite{Chen2020LearningOS, Liu2020FewShotOR} that equip classifiers with such an ability. Recognizing outliers might lead to more sample efficiency as the state abstractor can signal a reset to an environment when the agent is in bad and/or unrecoverable states. Nevertheless, although our state abstractor is trained using the in-distribution data from expert demonstrations, it still helps LEO-based agents perform well.








\bibliographystyle{IEEEtran}
\bibliography{refs}

\end{document}